\documentclass[letterpaper, 10 pt, conference]{ieeeconf}
\IEEEoverridecommandlockouts    
\usepackage[dvipsnames]{xcolor}
\usepackage{multirow}
\usepackage{pifont}

\definecolor{CommentBlue}{rgb}{0,0,1}

\newcommand{\rgbd}{\mbox{RGB-D}\ } 
\newcommand{\cmark}{\ding{51}}%
\newcommand{\xmark}{\ding{55}}%

\newif\ifshowhyperparams
 
\showhyperparamstrue

\ifshowhyperparams

\else

\fi

\usepackage{dirtree}

\def\tech#1{\small{\texttt{#1}}\normalsize}
\def\rgbd{\mbox{RGB-D} }
\def\rgb{RGB }

\definecolor{deepblue}{rgb}{0,0,0.5}
\definecolor{deepred}{rgb}{0.6,0,0}
\definecolor{deepgreen}{rgb}{0,0.5,0}

\usepackage{stmaryrd}
\usepackage{listings}

\newcommand\pythonstyle{%
  \lstset{
    language=Python,
    basicstyle=\ttfamily,
    morekeywords={self},              
    keywordstyle=\bfseries\color{deepblue},
    emph={MyClass,__init__},          
    emphstyle=\ttb\color{deepred},    
    stringstyle=\color{deepgreen},
    frame=tblr,                         
    showstringspaces=false,
  }%
}

\lstnewenvironment{python}[1][]
  {\pythonstyle\lstset{#1}}
  {}

\usepackage{hyperref}

\usepackage{graphics}           
\usepackage{times}              
\usepackage{amsmath}            
\usepackage{amssymb}            
\usepackage{graphicx}
\usepackage{algorithm}
\usepackage[noend]{algpseudocode}
\usepackage{booktabs}
\usepackage{color}

\usepackage[font=small]{caption}

\def\secref#1{Sec.~\ref{#1}}
\def\figref#1{Fig.~\ref{#1}}
\def\tabref#1{Tab.~\ref{#1}}
\def\eqref#1{Eq.~(\ref{#1})}

\makeatletter
\usepackage{xspace}
\DeclareRobustCommand\onedot{\futurelet\@let@token\@onedot}
\def\@onedot{\ifx\@let@token.\else.\null\fi\xspace}
 
\def\ie{i.e\onedot}

\def\etal{{et al}\onedot}
\makeatother

\def\etalcite#1{\etal~\cite{#1}}

\usepackage{array}
\newcolumntype{L}[1]{>{\raggedright\let\newline\\\arraybackslash\hspace{0pt}}m{#1}}
\newcolumntype{C}[1]{>{\centering\let\newline\\\arraybackslash\hspace{0pt}}m{#1}}
\newcolumntype{R}[1]{>{\raggedleft\let\newline\\\arraybackslash\hspace{0pt}}m{#1}}

\renewcommand{\b}[1]{\textbf{#1}}

\newcommand{\RR}{\mathbb{R}}

\newcommand{\norm}[1]{\lVert#1\lVert}

\renewcommand{\b}[1]{\mbox{\boldmath$#1$}}

\newcommand{\vv}[1]{{\b #1}} 

\title{\LARGE \bf A Dataset and Benchmark for Shape Completion\\of Fruits for Agricultural Robotics}

\renewcommand{\and}[0]{\hspace{0.8cm}}
\author{Federico Magistri, Thomas L{\"a}be, Elias Marks, Sumanth Nagulavancha, Yue Pan, Claus Smitt,\\ Lasse Klingbeil, Michael Halstead, Heiner Kuhlmann, Chris McCool, Jens Behley, Cyrill Stachniss \vspace{-1em}%
\thanks{All authors are with the Center for Robotics at the University of Bonn, Germany. Chris McCool and Cyrill Stachniss are additionally with the Lamarr Institute for Machine Learning and Artificial Intelligence, Germany. Cyrill Stachniss is furthermore with the Department of Engineering Science at the University of Oxford, UK.}%
	\thanks{This work has been funded by the Deutsche Forschungsgemeinschaft (DFG, German Research Foundation) under Germany's Excellence Strategy, EXC-2070 -- 390732324 -- PhenoRob and under STA~1051/5-1 within the FOR 5351~(AID4Crops).
  }%
}

\begin{document}
\maketitle
\thispagestyle{empty}
\pagestyle{empty}

\begin{abstract}
  As the world population is expected to reach 10 billion by 2050, our agricultural production system needs to double its productivity despite a decline of human workforce in the agricultural sector.
  Autonomous robotic systems are one promising pathway to increase productivity by taking over labor-intensive manual tasks like fruit picking. To be effective, such systems need to monitor and interact with plants and fruits precisely, which is challenging due to the cluttered nature of agricultural environments causing, for example, strong occlusions. Thus, being able to estimate the complete 3D shapes of objects in presence of occlusions is crucial for automating operations such as fruit harvesting.
  In this paper, we propose the first publicly available 3D shape completion dataset for agricultural vision systems.
  We provide an \rgbd dataset for estimating the 3D shape of fruits. Specifically, our dataset contains \rgbd frames of single sweet peppers in lab conditions but also in a commercial greenhouse. For each fruit, we additionally collected high-precision point clouds that we use as ground truth.
  For acquiring the ground truth shape, we developed a measuring process that allows us to record data of real sweet pepper plants, both in the lab and in the greenhouse with high precision, and determine the shape of the sensed fruits.
  We release our dataset, consisting of almost 7,000 \rgbd frames belonging to more than 100 different fruits.
  We provide segmented \rgbd frames, with camera intrinsics to easily obtain colored point clouds, together with the corresponding high-precision, occlusion-free point clouds obtained with a high-precision laser scanner. We additionally enable evaluation of shape completion approaches on a hidden test set through a public challenge on a benchmark server.
\end{abstract}

\section{Introduction}
\label{sec:intro}

Our agricultural production system needs to double its production of food, feed, fiber, and fuel to cope with an ever-growing population. A promising solution to support us in achieving this goal is the usage of autonomous systems that can continuously monitor the field, selectively harvest fruits and crops, or spot diseases early enough to avoid loss of yield. Such systems need to estimate important features of fruits and crops such as size, shape, and health status from perceived sensor data. In agricultural environments, this is extremely challenging due to their cluttered nature leading to plant and fruits being partially occluded.

In this paper, we aim to facilitate the development of agricultural vision systems in presence of such occlusions. Occlusions are a key issue in developing vision systems in agricultural environments. A correct handling of occlusions could benefit a diverse number of agricultural tasks such as high-throughput phenotyping, yield estimation and forecasting, and robotic harvesting.
\begin{figure}[t]
  \centering
  \includegraphics[width=0.85\linewidth]{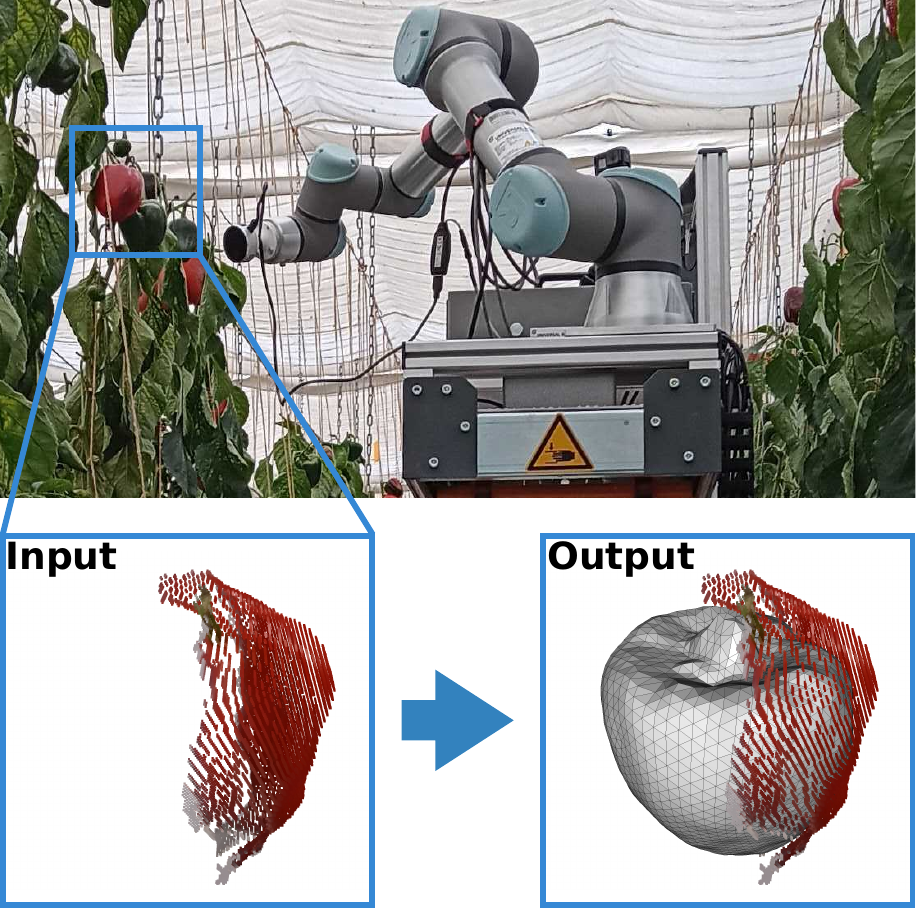}
  \caption{
    With our dataset, we tackle the problem of estimating the shape of fruits as a mesh (shown in grey) given a partial observation of an \rgbd sensor providing a colored point cloud.
    The estimated shape of the fruit is essential to allow safe grasping in situation with severe occlusions, like a greenhouse environment.
  }
  \label{fig:motivation}
\end{figure}

The main solutions for handling occlusions in agricultural environments have been constrained to 2D images, either by amodal instance segmentation methods that produce masks combining the visible and occluded part of crops~\cite{blok2021biosyseng} or by generative adversarial networks that estimate the appearance of crops as if there are no occlusions~\cite{kierdorf2022fai}. In contrast to the 2D case, fewer efforts have been made to tackle the occlusion problem in a 3D scene.
We are taking a different avenue to address occlusions in the context of fruit picking using 3D data provided by commonly employed \rgbd sensors.
In particular, manipulation tasks in robotics commonly employ 3D models of the objects to perform grasp prediction, but usually employ CAD models of known objects for this purpose, which cannot be pre-determined in an agricultural context, as each fruit has a different shape.
Therefore, we propose a dataset for promoting research on predicting a complete shape model of real-world fruits that can be used in manipulation scenarios~\cite{magistri2024ral}.
Thereby, we are addressing the challenging task of uncovering the full 3D shape of partially occluded fruits. Achieving this allows us to perform  more reliable fruit grasping, and, can be used to derive phenotypic traits about the fruit, such as size, shape, and quality.

The main contribution of this paper is the release of the first publicly available 3D shape completion dataset consisting of \rgbd frames and corresponding high-precision, occlusion-free point clouds obtained with a high-precision laser scanner.
The prediction task targets the estimation of the complete fruit shape from partial observations, as shown in \figref{fig:motivation}. 
In this paper, we also describe the process of generating the ground truth shape models for the sweet peppers in a real-world commercial greenhouse. We  specifically design a measuring process that allows us to acquire highly accurate and dense point cloud models of the sweet peppers.


In sum, our contributions can be summarized as follows:
(i) we release the first public dataset\footnote{
  Further details on downloading the dataset can be found at:\\
  \url{https://www.ipb.uni-bonn.de/data/shape_completion}} for 3D shape completion in agricultural environments,
(ii) we provide a CodaLab competition with a hidden test set to foster research in this area,
(iii) we provide a description of the process to obtain the dataset to facilitate other researcher's data collection efforts,
(iv) we release a Python-based development toolkit\footnote{Our development toolkit including a data loader is available at:\\
  \url{https://github.com/PRBonn/shape_completion_toolkit}} for handling the dataset and computing metrics.

\section{Related Work}
\label{sec:related}

Dataset and associated benchmarks have a long history in robotics and computer vision. 
They enable researchers to quantitatively and qualitatively measure the progress of research providing the foundation of a reproducible evaluation.
The datasets provide a starting point for investigation of novel research areas, where it is hard to acquire data or the data generation process requires specialized equipment.

In recent years, there has been an increasing interest in semantic interpretation of images in the agricultural context, both on arable crops~\cite{lottes2020jfr,weyler2022wacv,roggiolani2022icra} and horticulture~\cite{hani2020ral, sa2016sensors, barth2018cea}.
These works rely on large image datasets with pixel-accurate labels designed for various tasks: crop/weed segmentation~\cite{chebrolu2017ijrr,sa2018rs}, leaf segmentation~\cite{kierdorf2022jfr, weyler2024tpami}, fruit counting~\cite{smitt2021icra, perez2020cea}, and fruit size estimation~\cite{ferrer2023bioeng, gene2021pfuji}. We refer to Lu \etal~\cite{lu2020cea} for an overview of image datasets in agricultural environments. In contrast to these datasets, we provide 3D ground truth shapes of fruits in the agricultural context rather than segmentation labels.

While there are several image datasets available, few provide point cloud data and none of them tackle the occlusion problem in agriculture to the best of our knowledge. For instance, Chaudhury~\etal~\cite{chaudhury2020eccvws} proposed a synthetically generated point cloud dataset based on plant models with semantic labels. Furthermore, the Pheno4D dataset~\cite{schunck2021plosone} provides point clouds of tomato and maize plants captured with a high-precision laser scanner with labels of individual leaves that are consistent over time. James~\etalcite{james2024arxiv} provide a temporally consistent dataset of strawberry point clouds.
Dutagaci~\etal~\cite{dutagaci2020pm} release a rosebush plants 3D dataset acquired through X-ray tomography with semantic labels. Khanna~\etal~\cite{khanna2019pm} present a dataset containing color images, infra-red stereo image pairs, and multi-spectral camera images along with applied treatments and
weather conditions of the surroundings. Finally, Marks~\etal~\cite{marks2024arxiv} published a sugar beet point cloud dataset with leaves instance labels from real breeding plots. In contrast to these works, our dataset is specifically designed to tackle the occlusion problem in agricultural environments by providing complete shapes of sweet peppers with sub-millimeter accuracy.

Recently, a few datasets have been released for mapping applications in agricultural environments. The MAgro dataset~\cite{marzoa2023ijrr} consists of robotic sensor data, such as 3D LiDAR, data from an inertial measurement unit, and wheel encoders, gathered in apple and pear orchards with calibrated RTK GNSS to evaluate localization methods. The Bacchus dataset~\cite{polvara2023jfr} captures the whole canopy growth of a vineyard tailored for mapping and localization algorithms for long-term autonomous robotic operation. Instead, we designed our dataset to estimate precisely the 3D shape of individual fruits, which could be exploited for robotic grasping applications.

In summary, our dataset complements available point cloud datasets by providing measurements from a greenhouse using an \rgbd sensor with occlusions caused by leaves and other fruits. By providing the accurately measured shapes of the sensed fruits, we release a dataset for shape completion from partial observations and furthermore enable unbiased and reproducible evaluation on a hidden test set.

\section{Our Dataset}
\label{sec:main}
We propose a dataset of fruits collected in two scenarios. First, data collected in a lab environment for controlled experiments and, second, data collected in a commercial greenhouse showing the full complexity of the application scenario. In our benchmark, we use data collected in the lab for training set, while we use data collected in the greenhouse for testing set. The validation set contains data coming from both environments. The reason for this separation lies in the difficulty in collecting greenhouse data, having a procedure to collect data in the lab allows us to have a large training set.

As inputs in both scenarios, we choose to use \rgbd frames collected with an Intel RealSense d435i, which is a commonly used camera in agricultural robotics research. For collecting ground truth data capturing the shape of the fruits, we use a high-precision LiDAR, a Perceptron V5 laser scanner, mounted on a non-actuated measuring arm, a Hexagon ROMER Infinite 2.0 arm that provides sub-millimeter measurements. In this way, we can manually position the laser scanner to cover the fruits as good as possible. For more details on the LiDAR and the measuring arm, we refer to Schunck~\etalcite{schunck2021plosone}.
For both scenarios, we designed a specific measurement procedure to align the measurements from the \rgbd sensor with the accurate ground truth point cloud of the LiDAR such that we have a pose in the coordinate frame of the LiDAR.

The recording setup in the lab scenario allows us to register the different point clouds from the \rgb sensor and the LiDAR in a common reference frame. Here, we exploit a known environment structure to automatically determine the pose of the \rgbd sensor in the coordinate system of the LiDAR as described in \secref{sec:lab}.

In the greenhouse scenario, we have to adapt the measurement procedure as we cannot scan the fruits with the high-precision scanner on the trees, but still want to capture the plant in a natural environment including the occlusions caused by leaves and other parts of the plants.
Therefore, we designed a process that allows to associate the real scanned fruits with the measured fruit shapes of the ground truth point cloud, which we describe in \secref{sec:greenhouse}. Note that additional sensors such as an IMU can be used to improve the mapping accuracy. Despite having access to one within the camera, we did not use it as we found the mapping accuracy satisfactory.

\begin{figure}[t]
  \centering
  \includegraphics[width=0.65\linewidth]{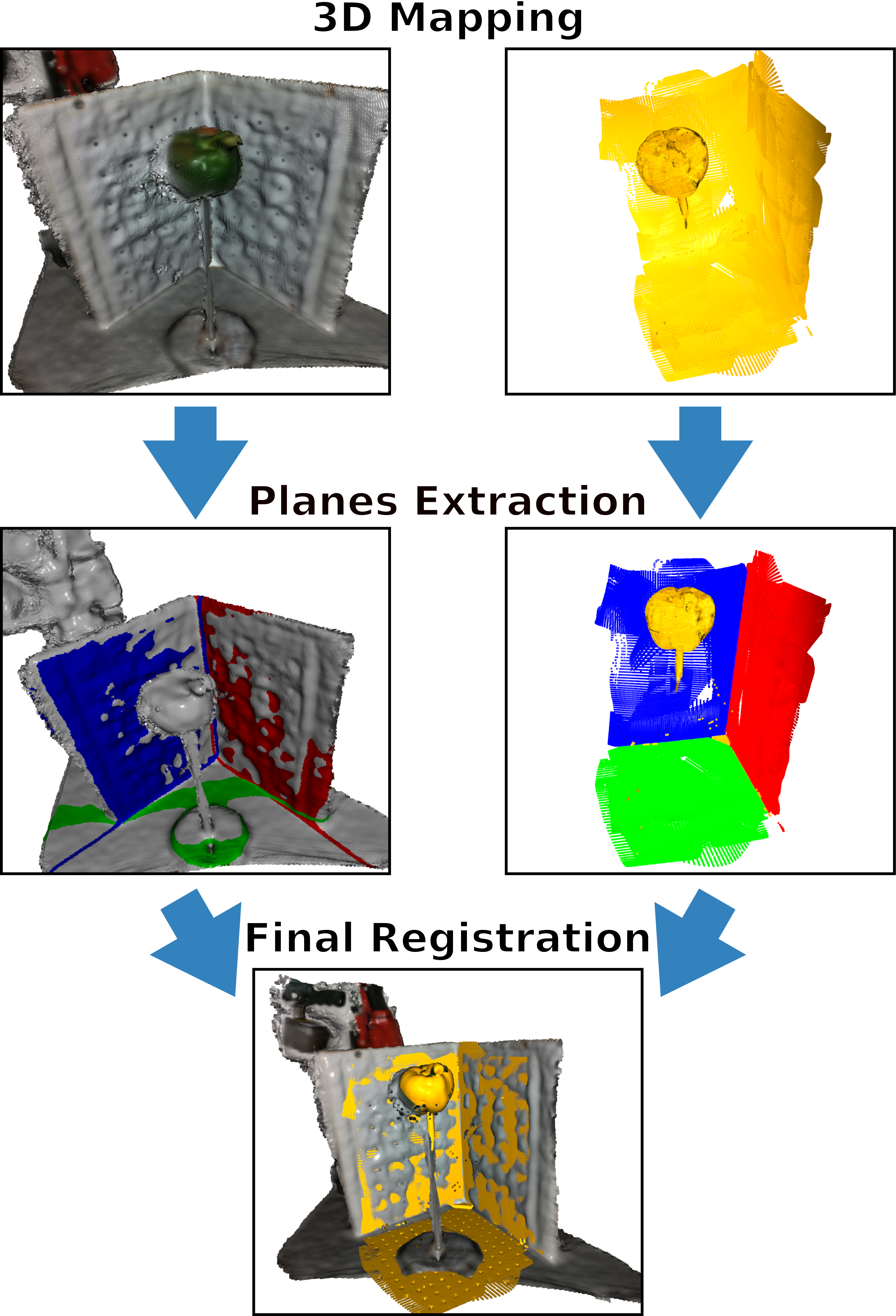}
  \caption{Registration procedure for the lab scenario. Given the TSDF-aligned \rgbd frames with the corresponding point cloud (on the left side) and the dense point cloud of the LiDAR (on the right side), we can estimate planes in each point cloud (shown in the middle). With an initial pose estimated via the extracted planes, we can register both point clouds using ICP automatically resulting in the final registration.
  }
  \vspace{-1em}
  \label{fig:labdata}
\end{figure}

\subsection{Lab Scenario}
\label{sec:lab}

In the lab scenario, we are able to record data of different fruits from more viewpoints compared to the greenhouse scenario. For each fruit, we collect the \rgbd frames by manually moving the camera around the fruit and trying to obtain as many viewpoints as possible. Afterwards, we repeat the process with the LiDAR covering the complete fruit. We want to highlight that it was not possible to scan the fruit with the different sensors at the same time, since the camera has a minimum range of $~30\,\text{cm}$ while the LiDAR operates at a closer range, \ie, $10\,\text{cm}$ on average.

Our goal is twofold. First, we want to use the laser scanner data from the test set as ground truth, while using the \rgbd data as input. Second, we want to use the laser scanner data to allow learning shape priors from high-quality point clouds that can be helpful when working on noisy point clouds obtained with a \rgbd camera. To achieve this, we need to register the data coming from the two sensors into a single coordinate system. A high-level overview of the registration process is shown in \figref{fig:labdata}.

To register the sensor data, we start by registering each \rgbd image belonging to the same fruit to each other with a standard TSDF fusion pipeline~\cite{curless1996siggraph, newcombe2011ismar, zhou2018arxiv}. This provides us with locally consistent poses of each recorded \rgbd image. The results of this step are the camera poses for each \rgbd frame in a local coordinate system of the first \rgbd image and a mesh of the object extracted via marching cubes~\cite{lorensen1987siggraph}.
To register the point cloud of the LiDAR to the points of the mesh from the \rgbd sensor, we apply iterative closest point~(ICP)~\cite{besl1992pami}, where we approximate the initial transformation between the two point clouds using the environment structure. To initialize the transformation between the point clouds, we scan the fruits inside three perpendicular planes and exploit this structure in the registration.

To determine the initial transformation, we extract the planes using a RANSAC approach which determines the three most dominant planes. To match the planes, we simply assume that for the first \rgbd frame we are approximately looking at the intersection point of
the three planes.
This results in three pairs of planes with normals $(\b{n}_i,  \b{n}'_i), i=1,2,3$, where we have normals $\b{n}_i \in \RR^{3}$ from the planes in the first point cloud and $\b{n}'_i \in \RR^{3}$ of the planes in the second point cloud.
The rotation matrix $\b{R} \in \RR^{3\times 3}$ between the two sets with known associations between the normals $(\b{n}_i,  \b{n}'_i)$ can be calculated
as follows~\cite{foerstner2016pcvbook}:
\begin{align}
  \medmuskip=1mu
  \b{R} & =  \frac{\b{n}'_1 (\b{n}_2 \times \b{n}_3)^\top  +  \b{n}'_2 (\b{n}_3 \times \b{n}_1)^\top  +  \b{n}'_3 (\b{n}_1 \times \b{n}_2)^\top}{ \det ( \b{n}_1^\top  \quad \b{n}_2^\top \quad \b{n}_3^\top  )},
\end{align}
where $\det(\cdot)$ corresponds to the determinant of the matrix $( \b{n}_1^\top  \quad \b{n}_2^\top \quad \b{n}_3^\top  )$ with the normals $\b{n}_i$  as column vectors.
The  matrix $\b{R}$ is usually not an orthonormal matrix because of the measurement noise, but a true rotation matrix, can be computed using a singular value decomposition of $\b{R}$.

The translation between the two sets of planes is the difference between the intersection point of the planes in the \rgbd point cloud and the intersection point of the planes in the LiDAR point cloud.
After a fine-registration using ICP, we can compute the pose of every \rgbd frame in the coordinate system of the LiDAR point cloud.

\begin{figure*}[t]
  \centering
  \includegraphics[width=0.9\linewidth]{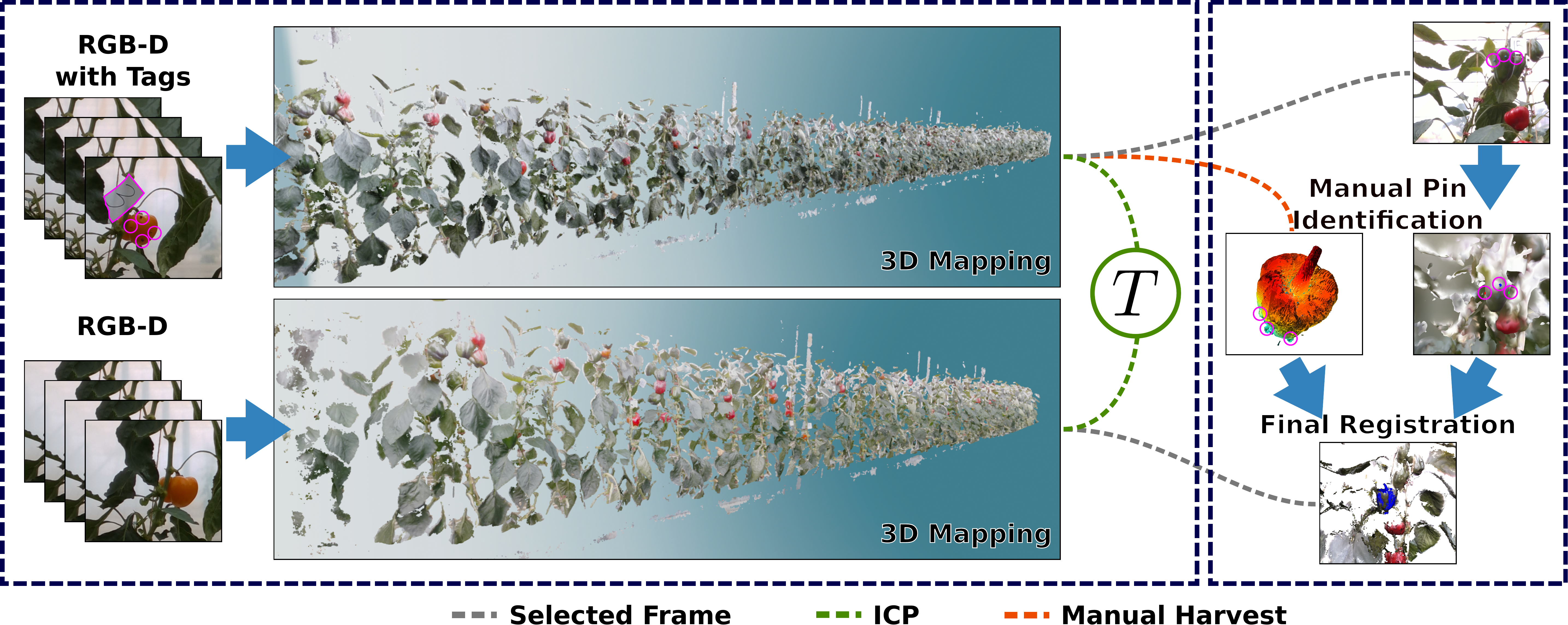}
  \caption{Measuring procedure to align greenhouse \rgbd frames with the corresponding ground truth point cloud generated by the LiDAR. Given two recordings with and without markers, we first align the photogrammetric point clouds via a transformation~$\b{T}$, which allows us to associate the sweet peppers without and with markers. Based on the markers and manually identified pins, we are able to determine the transformation of the scanned fruit in the \rgbd frame yielding the final registration.
  }
  \label{fig:ckadata}
\end{figure*}

\subsection{Greenhouse Scenario}
\label{sec:greenhouse}
For the greenhouse scenario, we collected data in a greenhouse at Campus Klein-Altendorf near Bonn, Germany, which we provide as validation and test sets. As before, the collected data is comprised of measurements of two different sensors. First, \rgbd frames collected with an autonomous robot equipped with three Intel RealSense d435i cameras mounted on a robotic platform as described by Smitt~\etalcite{smitt2021icra} and, second, high-precision scans of individual sweet peppers obtained with the aforementioned LiDAR as ground truth measurement.

In the lab scenario, we exploit the environment structure for aligning the data from different sensors. However, the same procedure is not possible as the sensors are located in different places. To register the individual sweet peppers in the \rgbd frames with their respective ground truth, we rely on a different measurement procedure.

We first perform a data collection run without any markers to get data as seen by a robot operating in the greenhouse.
For getting the ground truth shape via high-precision point cloud using the LiDAR, we augment the environment with markers to associate the harvested fruits with the already collected data. See \figref{fig:ckadata} for a visual impression of our greenhouse scenario and the measurement procedure.

More specifically, we manually tag individual sweet peppers in two ways: first, we place a label with an ID close to each fruit and, second, we insert multiple 3D markers, similar to paper pins, on each sweet pepper.
The ID ensures that we can identify the fruits after harvesting and scanning them with the LiDAR. The attached 3D markers allow us to register the scanned fruits in the \rgbd frames. After augmentation with markers, we deploy our robot again to collect data with the markers attached to the sweet peppers.

After these procedures we obtain: (i) a clean set of \rgbd frames that will be used as inputs for the shape completion challenge, (ii) a tagged set of \rgbd frames that we use for registration purposes, and (iii) high-precision point clouds for each sweet pepper that represents our ground truth for each individual fruit.
To register each fruit in the \rgbd frame they appear, we first conduct a 3D reconstruction for the complete plant row.
The TSDF fusion pipeline of the \rgbd frames that we used for the lab scenario did not yield convincing results, due to larger noise in the depth measurements under natural lighting conditions. To solve such an issue, we opted for an estimation of camera poses via an \rgb bundle adjustment using a commercial software for photogrammetric 3D reconstruction~\cite{Agisoft2024}. Together with the estimated camera poses, we can further derive a dense point cloud and a 3D mesh.

We scaled the two reconstructions, \ie, reconstruction of the first clean recording run and the recording with the markers, to the real scene scale using the data from the wheel odometry of the robot. As we need the relative transformation between these two 3D reconstructions for transforming fruit poses from the tagged dataset to the original dataset, we apply ICP to the dense point clouds for alignment. The initial transformation for ICP is estimated by manually identifying a single pair of frames of the two recording runs which show the same part of the row and computing their relative pose.

For aligning a point cloud of a fruit from the LiDAR to a frame of the tagged dataset, we manually measure three corresponding 3D markers in the laser scan and on the image frame.
Then we intersect the viewing rays constructed with the measured image coordinates with the reconstructed 3D mesh to get the 3D coordinates of the 3D markers in the tagged mesh.
Given these 3D point correspondences, we can estimate the relative pose of the scanned fruit in the tagged dataset. As we know the transformation to the
original dataset, we concatenate all transformations such that we end up in the pose of the scanned fruit in the \rgbd camera frame.
Finally, we apply ICP between the 3D mesh and a single scanned fruit to improve the pose estimate further.

With this, we obtain the ground truth shape of each sweet pepper and also ground truth poses with respect to the fruit canonical pose, \ie, with the peduncle pointing upwards.

\subsection{Curating the Benchmark Dataset}
\label{sec:dataset_preparation}

The objective of our dataset is to foster research in 3D shape completion and reconstruction in agricultural environments.
To simplify the task, we concentrate on the shape reconstruction and completion task for a segmented fruit, thus, factor out the process of generating an instance segmentation of a complete \rgbd frame and the pose estimation of the fruits.

To this end, we employed an instance segmentation approach trained on sweet peppers~\cite{smitt2021icra} to extract the parts of the \rgbd frame  \mbox{$\mathbf{I} \in \RR^{H \times W \times 4}$}, where each pixel $\mathbf{p}_i = (u,v)$ of the image frame of width $W$ and height $H$ corresponds to an RGB value $(r_i, g_i, b_i)$ and it's corresponding metric depth $d_i \in \RR$, that corresponds to the fruit $\mathcal{F}_j$, \ie, $\mathcal{F}_j = \{ \mathbf{p}_i | \mathbf{p}_i \in \mathbf{S}_j\}$, where $\mathbf{S}_j \in \{0,1\}^{H\times W}$ is the binary mask corresponding to the $j$-th fruit inside the image $\mathbf{I}$.

For each fruit point cloud $\mathcal{F}_j$ in the \rgbd point cloud, we then apply the know transformation from the \rgbd image to the LiDAR point cloud to transform the points of $\mathcal{F}_j$ into the LiDAR's canonical pose, \ie, with the peduncle pointing upwards.
Thus, for each fruit we provide: (i) the original \rgbd data,~$\mathcal{P}_j$, and (ii) the corresponding complete ground truth point cloud~$\hat{\mathcal{P}}_j$ obtained with the LiDAR.

\begin{figure}[t] 
  \centering
  \includegraphics[width=0.45\linewidth]{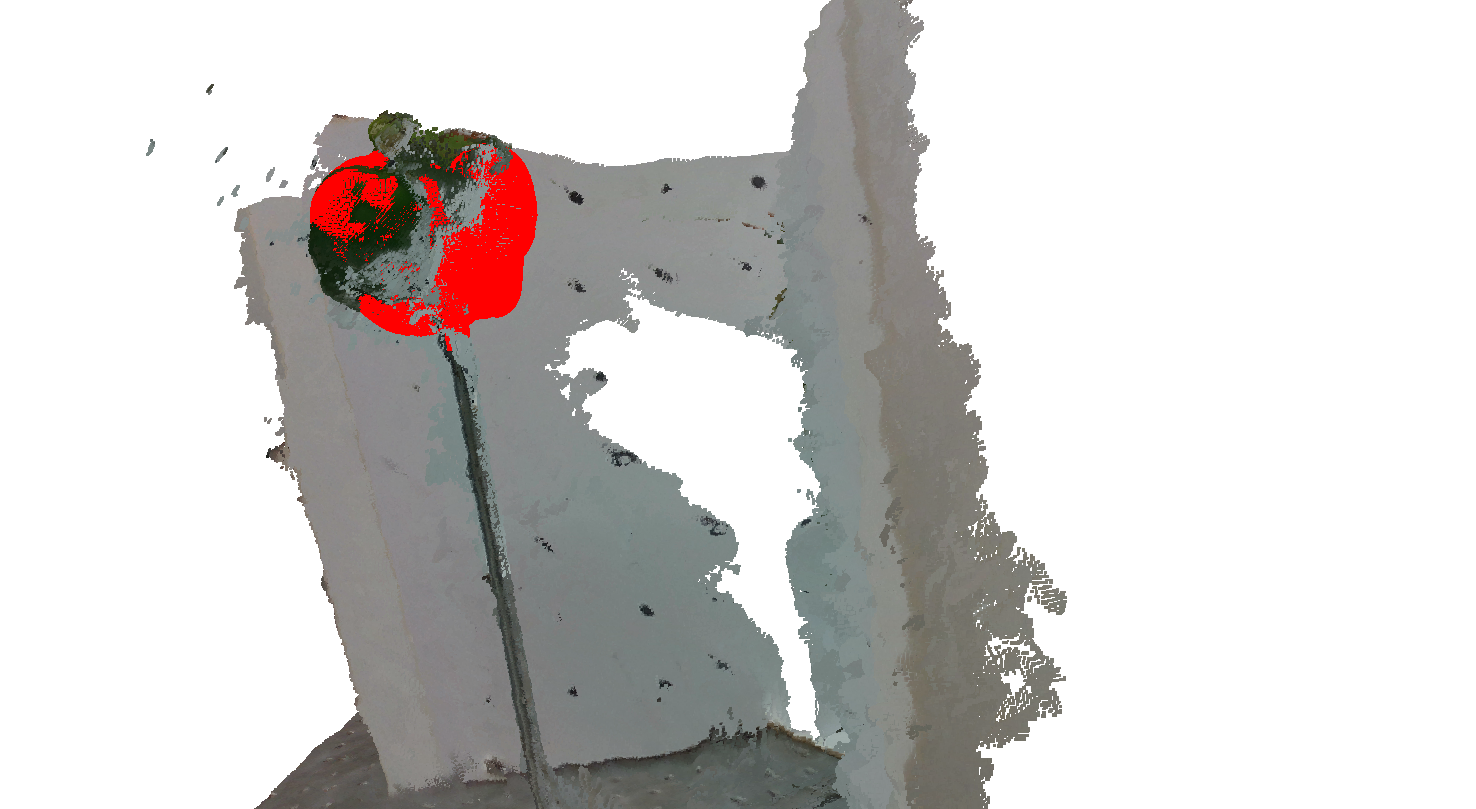} \quad
  \includegraphics[width=0.45\linewidth]{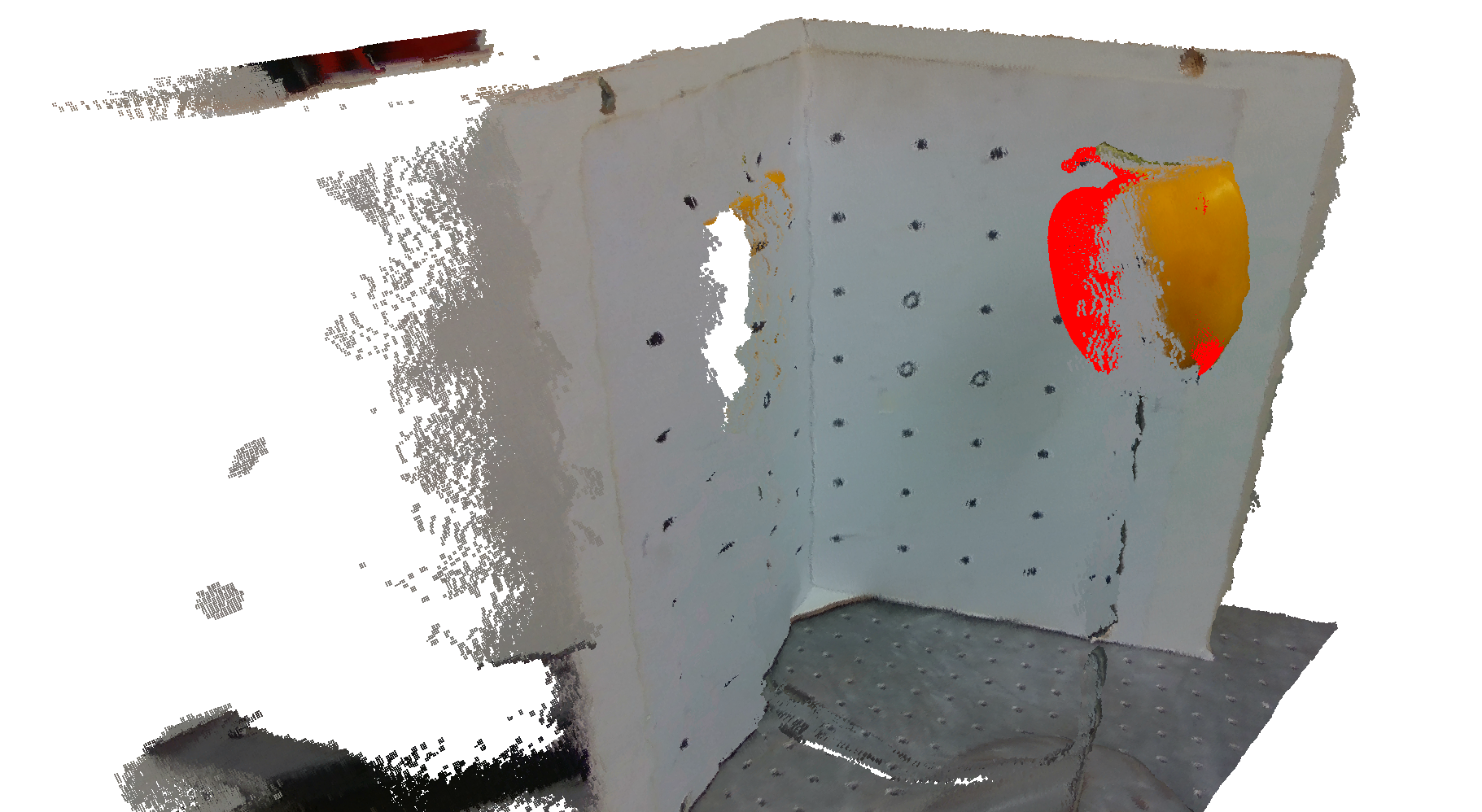} \\
  \vspace{1em}
  \includegraphics[width=0.45\linewidth]{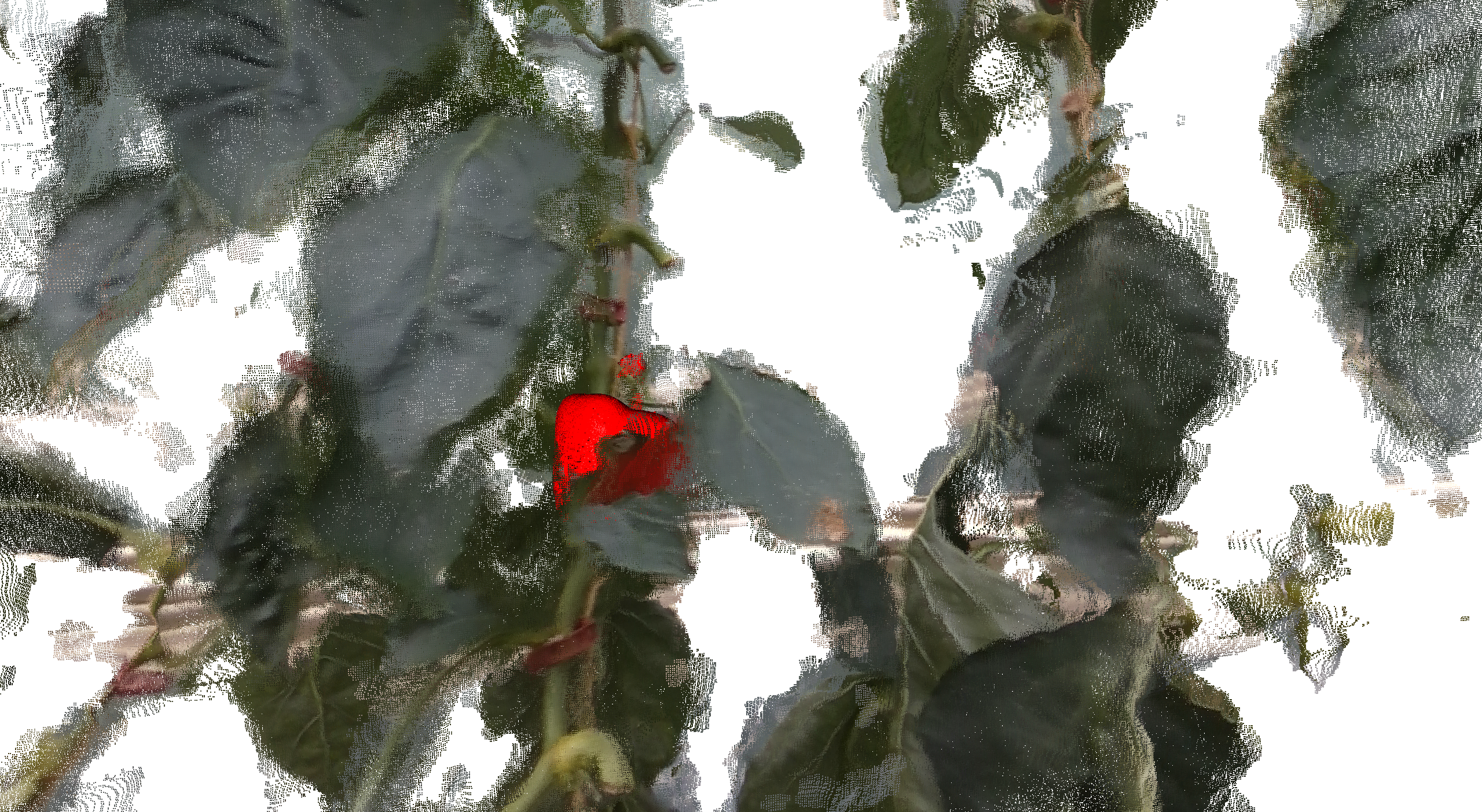} \quad
  \includegraphics[width=0.45\linewidth]{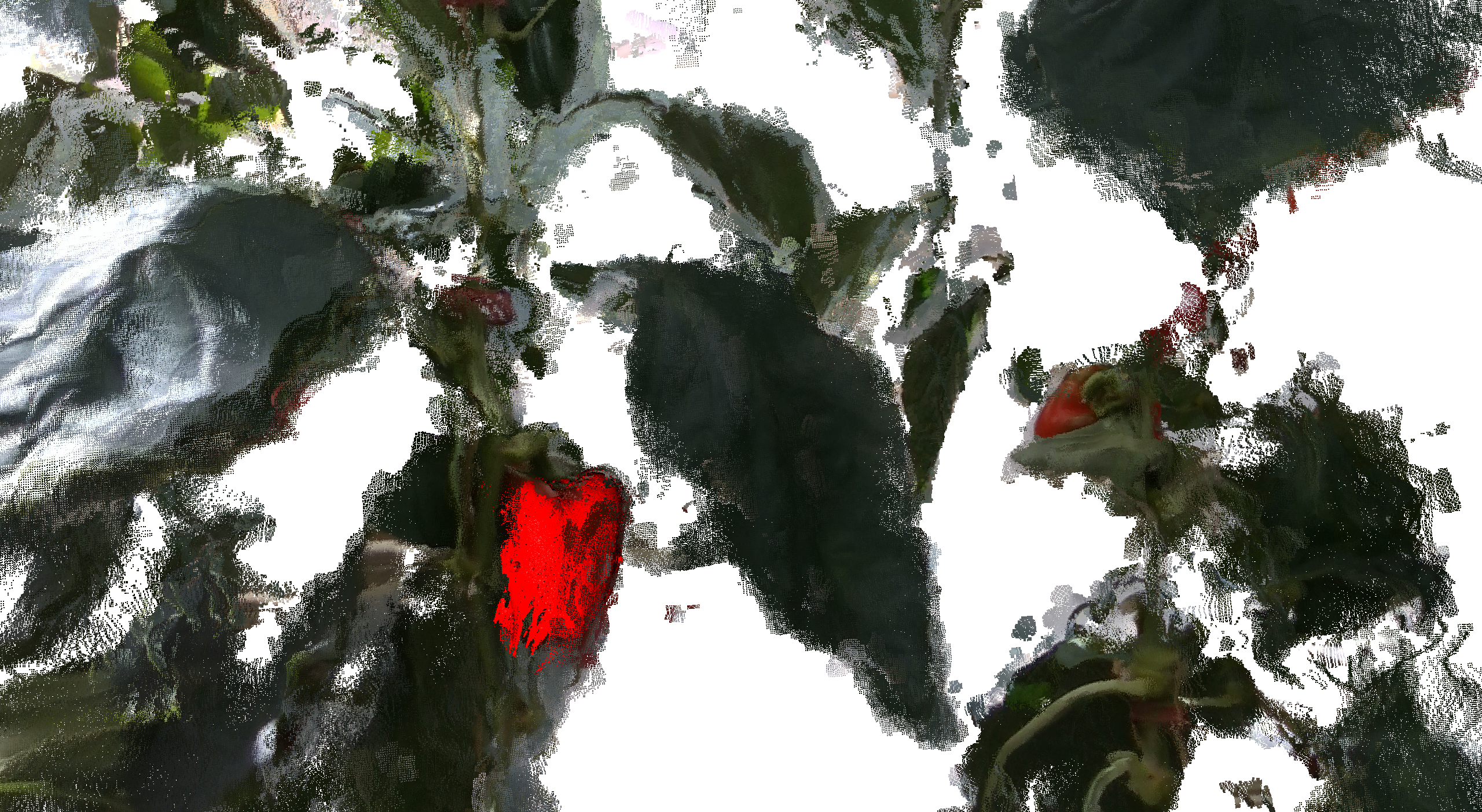} \\ 
  \caption{Qualitative example of our registation results in the lab (top) and greenhouse (bottom). Where we show the ground truth point cloud, in red, aligned with the corresponding \rgbd frames.}
  \vspace{-1em}
  \label{fig:data}
\end{figure}

For reproducible experimental evaluation in our benchmark, we split the data of the two scenarios into train, validation, and test set, where we ensured that fruits in the validation and test set are spatially separated. We want to highlight that greenhouse data is not present in the training set. Such a setting introduces a domain gap enabling us to evalute generalization capabilities of tested algorithms.
\tabref{tab:dataset_statistics} provides a summary of the provided data for the task of shape completion of sweet peppers.
Note that it is possible that the same fruit is visible from multiple viewpoints but share the same ground truth point cloud.
We refer to \figref{fig:data} for an qualitative impression of the registration results and refer to our dataset website\footnote{Our dataset website provides further details on the provided data: \url{https://www.ipb.uni-bonn.de/data/shape_completion}} for further details on the dataset organization.

\begin{table}[t]
  \centering
  \caption{Statistics of the provided data. L indicates data collected in the lab, G in the greenhouse.}
  \begin{tabular}{ccccc}
    \toprule
    Split      & \#Images & \#Fruits & Ground truth & Scenario \\
    \midrule
    Training   & 4580     & 66       & \cmark       & L        \\
    Validation & 1387     & 25       & \cmark       & L + G    \\
    Test       & 980      & 38       & \xmark       & G        \\
    \bottomrule
  \end{tabular}
  \label{tab:dataset_statistics}
\end{table}

\section{Shape Completion Benchmark}

Together with the dataset, we provide a benchmark for the reproducible and unbiased evaluation of shape completion approaches in the agricultural domain. Therefore, we concentrate here on the shape completion in a setting of reconstructing the fruit in a canonical orientation, where we factor out the process of determining the location of the fruits and provide \rgbd information in canonical fruit-specific local reference frame as described in \secref{sec:dataset_preparation}.
\begin{figure}[t]
  \centering
  \includegraphics[width=0.75\linewidth]{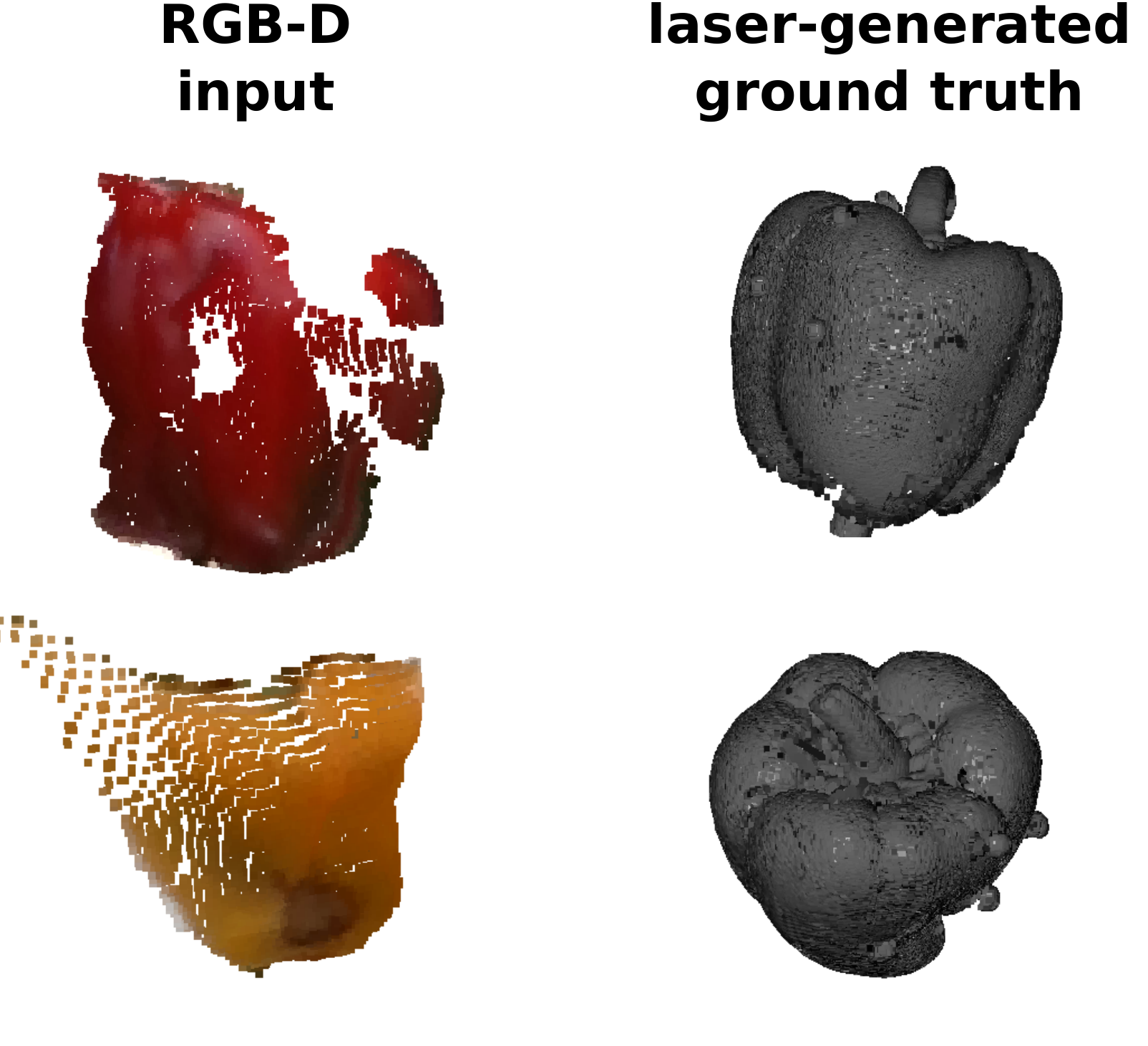}
  \caption{We show few example of input point clouds (left) and ground truth point clouds (right). The shape completion tasks involve estimating a complete 3D mesh from a partial and noisy point cloud.}
  \vspace{-1em}
  \label{fig:input-gt}
\end{figure}
In our benchmark, we expect the estimation of a complete 3D mesh $\mathcal{M}$ given partial observation of fruits $\mathcal{P}$ provided by an \rgbd sensor. The desired mesh $\mathcal{M} = (\mathcal{V}, \mathcal{T})$, represented by $N$ vertices $\mathcal{V} = \{\mathbf{v}_1, \dots, \mathbf{v}_N\}$ and $M$ triangles $\mathcal{T} = \{(i,j,k) | i, j, k \in \{1, \dots, N\} \}$ defined as tuples with the corresponding vertex indexes, should represent a closed surface of the complete fruit.
Thus, the task can be described as the problem of finding the complete mesh $\mathcal{M}$ of a given colored point cloud $\mathcal{P}$ in a canonical frame, see~\figref{fig:input-gt}.

\subsection{Metrics}

We report four different metrics: f-score, precision, recall, and Chamfer distance for evaluating 3D shapes, but rely on
the f-score integrating precision and recall as our main metric.
Precision, recall, and Chamfer distance are evaluated on point cloud $\mathcal{R}$ generated by densely sampling of the reconstructed mesh $\mathcal{M}$ and the corresponding point cloud  $\mathcal{G}$ of the LiDAR.

In line with Knapitsch \etalcite{knapitsch2017tog}, we compute the precision $p(\rho)$, \ie, which is the proportion of the points of the reconstruction $\mathcal{R}$ that are close to the ground truth $\mathcal{G}$ and recall $r(\rho)$, \ie, the proportion of the point of the ground truth $\mathcal{G}$ are close to the reconstructed mesh point cloud $\mathcal{R}$, where closeness is given by a threshold distance $\rho$:
\begin{align}
  p(\rho) & = \frac{100}{|\mathcal{R}|}\sum_{\vv{r} \in \mathcal{R}} \left\llbracket \mathop{\rm min}_{\vv{g} \in \mathcal{G}} || \vv{r} - \vv{g} || < \rho  \right\rrbracket, \\
  r(\rho) & = \frac{100}{|\mathcal{G}|}\sum_{\vv{g} \in \mathcal{G}} \left\llbracket \mathop{\rm min}_{\vv{r} \in \mathcal{R}} || \vv{g} - \vv{r} || < \rho  \right\rrbracket,
  \label{eq:precision_recall}
\end{align}
where $\vv g \in \RR^3$ and $\vv r \in \RR^3$ are points from $\mathcal{G}$ and $\mathcal{R}$, respectively.
Here, the operator~$\llbracket c \rrbracket$ is the Iverson bracket, \ie, if the condition $c$ within the brackets is satisfied it evaluates to $1$, otherwise to $0$.
Rather than choosing a fixed threshold $\rho$ like Knapitsch \etal \cite{knapitsch2017tog}, we evaluate precision and recall at different \mbox{$\rho$ values, \ie, $\rho \in \{0.01\,\text{m}, 0.02\,\text{m}, \dots ,0.1\,\text{m}\}$} with interval $\Delta = 0.01\,\text{m}$.
We compute the area under the curve for the precision $\bar{p}$ and recall $\bar{r}$ as follows:
\begin{align}
  \bar{p} & = \frac{1}{\eta} \sum_{\rho \in \{0.01\,\text{m}, 0.02\,\text{m}, \dots ,0.1\,\text{m}\}} \Delta\, p(\rho)  \\
  \bar{r} & = \frac{1}{\eta} \sum_{\rho \in \{0.01\,\text{m}, 0.02\,\text{m}, \dots ,0.1\,\text{m}\}} \Delta\, r(\rho),
  \label{eq:area_under_curve}
\end{align}
where $\eta$ is a normalizer that corresponds to the precision or recall of a perfect estimator evaluated with aforementioned thresholds to arrive at $\bar{p} \in [0,1]$ and $\bar{r} \in [0,1]$.

Finally, the f-score $f$ is given by:
\begin{align}
  f & = \frac{2\,\bar{p} \, \bar{r}}{\bar{p}+\bar{r}}.
\end{align}


We additionally evaluate the Chamfer distance $D_\text{C}$. It is the average symmetric squared distance $\bar{d}^2$ of each point to its nearest neighbor in the other point cloud:
\begin{align}
  D_\text{C}(\mathcal{G},\mathcal{R}) & = \frac{\bar{d}^2(\mathcal{G},\mathcal{R})}{2}  + \frac{\bar{d}^2(\mathcal{R},\mathcal{G})}{2},
\end{align}
with
\begin{align}
  \bar{d}^2(\mathcal{P}_i,\mathcal{P}_j) & = \frac{1}{|\mathcal{P}_i|} \sum_{\vv{x}_i \in \mathcal{P}_i} \min_{\vv{x}_j \in \mathcal{P}_j} \norm{\vv{x}_i-\vv{x}_j}_2^2.
\end{align}

For the ranking of the entries submitted to our competition, we use the F1-score as a primary metric and the Chamfer distance as secondary metric in case of a tie.

\subsection{CodaLab Competition}

To evaluate predictions on the hidden test set, we expect a specific directory structure in our CodaLab competition~\cite{pavao2023jmlr}. We ask the participants to submit a zip file that contains one \tech{.ply} file for each fruit id. Note that we expect the fruit to be in the canonical pose, \ie, with the peduncle pointing upwards, this is trivial to obtain given the provided pose for each \rgbd frame. More details can be found on the CodaLab submission page~\footnote{The CodaLab competition is available at: \url{https://codalab.lisn.upsaclay.fr/competitions/18987}}.

To speed up the usage of our dataset, we provide a development kit providing a PyTorch-based data loader and a small library for computing the metrics that we use in the CodaLab competition.

\subsection{Baselines}

To facilitate the comparisons with previous work, namely CoRe~\cite{magistri2022ral-iros}, HoMa~\cite{pan2023iros}, and T-CoRe~\cite{magistri2024icra}, we report the performances on the test set in~\tabref{tab:baseline_test}. Additionally, we report the performances on the validation set in~\tabref{tab:baseline_val} to enable comparison of novel approaches in ablation studies using the validation set.
From our experiments, T-CoRe provides better predictions on the test set reaching $5.06$\,mm of Chamfer distance and $58.0$\,\% of f-score. In contrast, HoMa is the best-performing method on the validation set, obtaining $4.2$\,mm of Chamfer distance and $65.7$\,\% of f-score. We believe this decrease in performance on the test set can be a symptom of overfitting on the recording conditions. Note that the validation set contains data from the lab and from the greenhouse, while the test set only contains samples collected in the greenhouse.

Finally, HoMa and T-CoRe yield better results than CoRe across all metrics, which is expected given that it is the only semi-supervised method. Notably, CoRe is the only approach operating on images, while T-CoRe and HoMa use point clouds. We believe this also contributes to the final performances.



\begin{table}[t]
 \centering
  \caption{Baseline results on the test set.  The $\downarrow$ and $\uparrow$ indicate that lower or higher values mean better performance.}
 \label{tab:baseline_test}
 \begin{tabular}{ccccc} 
   \toprule
   \multirow{2}{*}{\textbf{Approach}} & \textbf{$\boldsymbol{D}_\text{C}$} [mm] & \textbf{\textit{f}} [\%] & $\boldsymbol{\bar{p}}$ [\%] & $\boldsymbol{\bar{r}}$ [\%] \\
    & $\downarrow$ avg & $\uparrow$ avg & $\uparrow$ avg & $\uparrow$ avg \\
   \midrule
   CoRe~\cite{magistri2022ral-iros}           & 6.84 & 45.61 & 43.87 & 48.56 \\     
   HoMa~\cite{pan2023iros}                    & 5.39 &  57.11 & 55.63 & \b{58.88} \\
   T-CoRe~\cite{magistri2024icra}             & \b{5.06} & \b{58.04} & \b{58.60} & 57.76 \\     
   \bottomrule
 \end{tabular}
 \vspace{.5em}
\end{table}

\begin{table}[t]
 \centering
  \caption{Baseline results on the validation set.  The $\downarrow$ and $\uparrow$ indicate that lower or higher values mean better performance.}
 \label{tab:baseline_val}
 \begin{tabular}{ccccc} 
   \toprule
   \multirow{2}{*}{\textbf{Approach}} & \textbf{$\boldsymbol{D}_\text{C}$} [mm] & \textbf{\textit{f}} [\%] & $\boldsymbol{\bar{p}}$ [\%] & $\boldsymbol{\bar{r}}$ [\%] \\
    & $\downarrow$ avg & $\uparrow$ avg & $\uparrow$ avg & $\uparrow$ avg \\
   \midrule
   CoRe~\cite{magistri2022ral-iros}           & 7.18 & 42.48 & 38.92 & 47.52 \\     
   HoMa~\cite{pan2023iros}                    & \b{4.19} & \b{65.65} & \b{63.58} & \b{68.01} \\
   T-CoRe~\cite{magistri2024icra}             & 4.69 & 60.61 & 60.59 & 60.82 \\     
   \bottomrule
 \end{tabular}
 \vspace{-0.5em}
\end{table}

\section{Conclusion}
In this paper, we presented a novel dataset for 3D shape completion of sweet peppers in agricultural environments.
Our dataset consists of \rgbd frames collected in the lab and a commercial greenhouse, together with dense high-precision point clouds as ground truth of the sweet peppers, where we aligned the \rgbd data and the ground truth point cloud with a tailored measurement procedure.
We additionally provide a CodaLab competition with a hidden test set to ensure fair comparisons and automatic evaluations to foster research into agricultural vision systems.

Our competion has already been used by more than 50 participants with a substantial increase in performances~\cite{chen2024arxiv}. We believe this further demonstrates that the task used in the benchmark of this paper is not solved yet with plenty of room for improvements and at the same time the benefits of curating public benchmark with hidden test sets.

In future, we also plan to provide ground truth for other applications, such as 3D object pose estimation of fruits, where we leverage our measurement procedure to determine an accurate pose of visible sweet peppers based on the alignment between the marker point clouds and our ground truth measurements.


%
%
\bibliographystyle{abbrv}
\bibliography{glorified,new}

\begin{thebibliography}{10}

\bibitem{Agisoft2024}
{\relax Agisoft LLC}.
\newblock {Agisoft Metashape Professional}.
\newblock \url{https://www.agisoft.com/}.
\newblock Accessed: April 2024.

\bibitem{barth2018cea}
R.~Barth, J.~IJsselmuiden, J.~Hemming, and E.~J. Van~Henten.
\newblock {Data Synthesis Methods for Semantic Segmentation in Agriculture: A
  Capsicum Annuum Dataset}.
\newblock {\em Computers and Electronics in Agriculture}, 144:284--296, 2018.

\bibitem{besl1992pami}
P.~Besl and N.~McKay.
\newblock {A Method for Registration of 3D Shapes}.
\newblock {\em IEEE Trans.~on Pattern Analysis and Machine Intelligence
  (TPAMI)}, 14(2):239--256, 1992.

\bibitem{blok2021biosyseng}
P.~M. Blok, E.~J. van Henten, F.~K. van Evert, and G.~Kootstra.
\newblock Image-based size estimation of broccoli heads under varying degrees
  of occlusion.
\newblock {\em Biosystems Engineering}, 208:213--233, 2021.

\bibitem{chaudhury2020eccvws}
A.~Chaudhury, F.~Boudon, and C.~Godin.
\newblock {3D Plant Phenotyping: All You Need is Labelled Point Cloud Data}.
\newblock In {\em {ECCV Workshop on Computer Vision Problems in Plant
  Phenotyping}}, 2020.

\bibitem{chebrolu2017ijrr}
N.~Chebrolu, P.~Lottes, A.~Schaefer, W.~Winterhalter, W.~Burgard, and
  C.~Stachniss.
\newblock {Agricultural Robot Dataset for Plant Classification, Localization
  and Mapping on Sugar Beet Fields}.
\newblock {\em Intl.~Journal~of Robotics Research (IJRR)}, 36(10):1045--1052,
  2017.

\bibitem{chen2024arxiv}
Z.~Chen, T.~Wei, Z.~Zhao, J.~S. Lim, Y.~Luo, H.~Zhang, X.~Yu, S.~Chapman, and
  Z.~Huang.
\newblock Cf-prnet: Coarse-to-fine prototype refining network for point cloud
  completion and reconstruction.
\newblock {\em arXiv preprint}, arXiv:2409.08443, 2024.

\bibitem{curless1996siggraph}
B.~Curless and M.~Levoy.
\newblock {A Volumetric Method for Building Complex Models from Range Images}.
\newblock In {\em Proc.~of the Intl.~Conf.~on Computer Graphics and Interactive
  Techniques (SIGGRAPH)}, 1996.

\bibitem{dutagaci2020pm}
H.~Dutağacı, P.~Rasti, G.~Galopin, and D.~Rousseau.
\newblock {ROSE-X: an Annotated Data Set for Evaluation of 3D Plant Organ
  Segmentation Methods}.
\newblock {\em Plant Methods}, 16, 2020.

\bibitem{ferrer2023bioeng}
M.~Ferrer-Ferrer, J.~Ruiz-Hidalgo, E.~Gregorio, V.~Vilaplana, J.-R. Morros, and
  J.~Gen{\'e}-Mola.
\newblock {Simultaneous Fruit Detection and Size Estimation Using Multitask
  Deep Neural Networks}.
\newblock {\em Biosystems Engineering}, 233:63--75, 2023.

\bibitem{foerstner2016pcvbook}
W.~F{\"o}rstner and B.~Wrobel.
\newblock {\em {Photogrammetric Computer Vision -- Statistics, Geometry,
  Orientation and Reconstruction}}.
\newblock Springer Verlag, 2016.

\bibitem{gene2021pfuji}
J.~Gen{\'e}-Mola, R.~Sanz-Cortiella, J.~R. Rosell-Polo, A.~Escol{\`a}, and
  E.~Gregorio.
\newblock {PFuji-Size Dataset: A Collection of Images and
  Photogrammetry-Derived 3D Point Clouds with gruth Annotations for Fuji Apple
  Detection and Size Estimation in Field Conditions}.
\newblock {\em Data in Brief}, 39:107629, 2021.

\bibitem{hani2020ral}
N.~H{\"a}ni, P.~Roy, and V.~Isler.
\newblock {MinneApple: a Benchmark Dataset for Apple Detection and
  Segmentation}.
\newblock {\em IEEE Robotics and Automation Letters (RA-L)}, 5(2):852--858,
  2020.

\bibitem{james2024arxiv}
K.~M.~F. James, K.~Heiwolt, D.~J. Sargent, and G.~Cielniak.
\newblock {Lincoln's Annotated Spatio-Temporal Strawberry Dataset
  (LAST-Straw)}.
\newblock {\em arXiv preprint}, arXiv:2403.00566, 2024.

\bibitem{khanna2019pm}
R.~Khanna, L.~Schmid, A.~Walter, J.~Nieto, R.~Siegwart, and F.~Liebisch.
\newblock {A Spatio Temporal Spectral Framework for Plant Stress Phenotyping}.
\newblock {\em Plant Methods}, 15, 2019.

\bibitem{kierdorf2022jfr}
J.~Kierdorf, L.~V. Junker-Frohn, M.~Delaney, M.~D. Olave, A.~Burkart,
  H.~Jaenicke, O.~Muller, U.~Rascher, and R.~Roscher.
\newblock Growliflower: An image time series dataset for growth analysis of
  cauliflower.
\newblock {\em Journal of Field Robotics (JFR)}, 40(2):173--192, 2022.

\bibitem{kierdorf2022fai}
J.~Kierdorf, I.~Weber, A.~Kicherer, L.~Zabawa, L.~Drees, and R.~Roscher.
\newblock Behind the leaves: Estimation of occluded grapevine berries with
  conditional generative adversarial networks.
\newblock {\em Frontiers in Artificial Intelligence}, 5:830026, 2022.

\bibitem{knapitsch2017tog}
A.~Knapitsch, J.~Park, Q.~Zhou, and V.~Koltun.
\newblock Tanks and temples: Benchmarking large-scale scene reconstruction.
\newblock {\em ACM Trans.~on Graphics}, 36(4):1--13, 2017.

\bibitem{lorensen1987siggraph}
W.~Lorensen and H.~Cline.
\newblock {Marching Cubes: a High Resolution 3D Surface Construction
  Algorithm}.
\newblock In {\em Proc.~of the Intl.~Conf.~on Computer Graphics and Interactive
  Techniques (SIGGRAPH)}, pages 163--169, 1987.

\bibitem{lottes2020jfr}
P.~Lottes, J.~Behley, N.~Chebrolu, A.~Milioto, and C.~Stachniss.
\newblock {Robust Joint Stem Detection and Crop-Weed Classification using Image
  Sequences for Plant-Specific Treatment in Precision Farming}.
\newblock {\em Journal of Field Robotics (JFR)}, 37:20--34, 2020.

\bibitem{lu2020cea}
Y.~Lu and S.~Young.
\newblock {A Survey of Public Datasets for Computer Vision Tasks in Precision
  Agriculture}.
\newblock {\em Computers and Electronics in Agriculture}, 178:105760, 2020.

\bibitem{magistri2024icra}
F.~Magistri, R.~Marcuzzi, E.~Marks, M.~Sodano, J.~Behley, and C.~Stachniss.
\newblock {Efficient and Accurate Transformer-Based 3D Shape Completion and
  Reconstruction of Fruits for Agricultural Robots}.
\newblock In {\em Proc.~of the IEEE Intl.~Conf.~on Robotics \& Automation
  (ICRA)}, 2024.

\bibitem{magistri2022ral-iros}
F.~Magistri, E.~Marks, S.~Nagulavancha, I.~Vizzo, T.~L{\"a}be, J.~Behley,
  M.~Halstead, C.~McCool, and C.~Stachniss.
\newblock Contrastive 3d shape completion and reconstruction for agricultural
  robots using rgb-d frames.
\newblock {\em IEEE Robotics and Automation Letters (RA-L)}, 7(4):10120--10127,
  2022.

\bibitem{magistri2024ral}
F.~Magistri, Y.~Pan, J.~Bartels, J.~Behley, C.~Stachniss, and C.~Lehnert.
\newblock {Improving Robotic Fruit Harvesting Within Cluttered Environments
  Through 3D Shape Completion}.
\newblock {\em IEEE Robotics and Automation Letters (RA-L)}, 9(8):7357--7364,
  2024.

\bibitem{marks2024arxiv}
E.~Marks, J.~B{\"o}mer, F.~Magistri, A.~Sah, J.~Behley, and C.~Stachniss.
\newblock {BonnBeetClouds3D: A Dataset Towards Point Cloud-based Organ-level
  Phenotyping of Sugar Beet Plants under Field Conditions}.
\newblock In {\em Proc.~of the IEEE/RSJ Intl.~Conf.~on Intelligent Robots and
  Systems (IROS)}, 2023.

\bibitem{marzoa2023ijrr}
M.~Marzoa~Tanco, G.~Trinidad~Barnech, F.~Andrade, J.~Baliosian, M.~LLofriu,
  J.~Di~Martino, and G.~Tejera.
\newblock {MAgro dataset: A Dataset for Simultaneous Localization and Mapping
  in Agricultural Environments}.
\newblock {\em Intl.~Journal~of Robotics Research (IJRR)}, 43(5):591--601,
  2024.

\bibitem{newcombe2011ismar}
R.~A. Newcombe, S.~Izadi, O.~Hilliges, D.~Molyneaux, D.~Kim, A.~J. Davison,
  P.~Kohli, J.~Shotton, S.~Hodges, and A.~Fitzgibbon.
\newblock {KinectFusion: Real-Time Dense Surface Mapping and Tracking}.
\newblock In {\em Proc.~of the Intl.~Symposium~on Mixed and Augmented Reality
  (ISMAR)}, 2011.

\bibitem{pan2023iros}
Y.~Pan, F.~Magistri, T.~L\"abe, E.~Marks, C.~Smitt, C.~McCool, J.~Behley, and
  C.~Stachniss.
\newblock {Panoptic Mapping with Fruit Completion and Pose Estimation for
  Horticultural Robots}.
\newblock In {\em Proc.~of the IEEE/RSJ Intl.~Conf.~on Intelligent Robots and
  Systems (IROS)}, 2023.

\bibitem{pavao2023jmlr}
A.~Pavao, I.~Guyon, A.-C. Letournel, D.-T. Tran, X.~Baro, H.~J. Escalante,
  S.~Escalera, T.~Thomas, and Z.~Xu.
\newblock {CodaLab Competitions: An Open Source Platform to Organize Scientific
  Challenges}.
\newblock {\em Journal on Machine Learning Research~(JMLR)}, 24(198):1--6,
  2023.

\bibitem{polvara2023jfr}
R.~Polvara, S.~Molina, I.~Hroob, A.~Papadimitriou, K.~Tsiolis, D.~Giakoumis,
  S.~Likothanassis, D.~Tzovaras, G.~Cielniak, and M.~Hanheide.
\newblock {Bacchus Long-Term (BLT) Data Set: Acquisition of the Agricultural
  Multimodal BLT Data Set with Automated Robot Deployment}.
\newblock {\em Journal of Field Robotics (JFR)}, 2023.

\bibitem{perez2020cea}
I.~Pérez-Borrero, D.~Marín-Santos, M.~E. Gegúndez-Arias, and
  E.~Cortés-Ancos.
\newblock {A Fast and Accurate Deep Learning Method for Strawberry Instance
  Segmentation}.
\newblock {\em Computers and Electronics in Agriculture}, 178:105736, 2020.

\bibitem{roggiolani2022icra}
G.~Roggiolani, M.~Sodano, T.~Guadagnino, F.~Magistri, J.~Behley, and
  C.~Stachniss.
\newblock {Hierarchical Approach for Joint Semantic, Plant Instance, and Leaf
  Instance Segmentation in the Agricultural Domain}.
\newblock {\em Proc.~of the IEEE Intl.~Conf.~on Robotics \& Automation (ICRA)},
  2023.

\bibitem{sa2016sensors}
I.~Sa, Z.~Ge, F.~Dayoub, B.~Upcroft, T.~Perez, and C.~McCool.
\newblock Deepfruits: A fruit detection system using deep neural networks.
\newblock {\em Sensors}, 16(8):1222, 2016.

\bibitem{sa2018rs}
I.~Sa, M.~Popovic, R.~Khanna, Z.~Chen, P.~Lottes, F.~Liebisch, J.~Nieto,
  C.~Stachniss, and R.~Siegwart.
\newblock {WeedMap: A Large-Scale Semantic Weed Mapping Framework Using Aerial
  Multispectral Imaging and Deep Neural Network for Precision Farming}.
\newblock {\em Remote Sensing}, 10, 2018.

\bibitem{schunck2021plosone}
D.~Schunck, F.~Magistri, R.~Rosu, A.~Corneli{\ss}en, N.~Chebrolu, S.~Paulus,
  J.~L\'eon, S.~Behnke, C.~Stachniss, H.~Kuhlmann, and L.~Klingbeil.
\newblock {Pheno4D: A spatio-temporal dataset of maize and tomato plant point
  clouds for phenotyping and advanced plant analysis }.
\newblock {\em PLOS ONE}, 16(8):1--18, 2021.

\bibitem{smitt2021icra}
C.~Smitt, M.~Halstead, T.~Zaenker, M.~Bennewitz, and C.~McCool.
\newblock {PATHoBot}: {A} robot for glasshouse crop phenotyping and
  intervention.
\newblock In {\em Proc.~of the IEEE Intl.~Conf.~on Robotics \& Automation
  (ICRA)}, 2021.

\bibitem{weyler2024tpami}
J.~Weyler, F.~Magistri, E.~Marks, Y.~Chong, M.~Sodano, G.~Roggiolani,
  N.~Chebrolu, C.~Stachniss, and J.~Behley.
\newblock {PhenoBench: A Large Dataset and Benchmarks for Semantic Image
  Interpretation in the Agricultural Domain}.
\newblock {\em IEEE Trans.~on Pattern Analysis and Machine Intelligence
  (TPAMI)}, 46(12):9583--9594, 2024.

\bibitem{weyler2022wacv}
J.~Weyler, F.~Magistri, P.~Seitz, J.~Behley, and C.~Stachniss.
\newblock {In-Field Phenotyping Based on Crop Leaf and Plant Instance
  Segmentation}.
\newblock In {\em Proc.~of the IEEE Winter Conf.~on Applications of Computer
  Vision (WACV)}, 2022.

\bibitem{zhou2018arxiv}
Q.~Zhou, J.~Park, and V.~Koltun.
\newblock {Open3D}: {A} modern library for {3D} data processing.
\newblock {\em arXiv:1801.09847}, 2018.

\end{thebibliography}

\end{document}